\documentclass[letterpaper, 10 pt, conference]{ieeeconf}

\IEEEoverridecommandlockouts

\usepackage{epsfig} % for postscript graphics files
\usepackage{url}

\usepackage{amsmath} % assumes amsmath package installed
\usepackage{amssymb}  % assumes amsmath package installed
\usepackage{multirow}

\let\labelindent\relax
\usepackage{enumitem}% http://ctan.org/pkg/enumerate

\usepackage{multicol}        % used for the two-column index
\usepackage[bottom]{footmisc}% places footnotes at page bottom
\usepackage{algorithm}
\usepackage{algorithmic}
\usepackage{epstopdf}
\usepackage{subfig}
\usepackage{xcolor}

\makeatletter
\let\NAT@parse\undefined
\makeatother
\usepackage{hyperref}

\graphicspath{{figures/}}
\title{\LARGE \bf Functional Object-Oriented Network: Construction \& Expansion}

\author{David Paulius, Ahmad B. Jelodar and Yu Sun
\thanks{David Paulius, Ahmad B. Jelodar, and Yu Sun are with the Department of Computer Science and Engineering at the University of South Florida and members of the Robot Perception and Action Lab.
 \newline(Contact email: \{davidpaulius,yusun\}@mail.usf.edu)}
}

\begin{document}

\maketitle

\thispagestyle{empty}

\pagestyle{empty}

\begin{abstract}
We build upon the \emph{functional object-oriented network} (FOON), a structured knowledge representation which is constructed from observations of human activities and manipulations.
A FOON can be used for representing object-motion affordances.
Knowledge retrieval through graph search allows us to obtain novel manipulation sequences using knowledge spanning across many video sources, hence the novelty in our approach.
However, we are limited to the sources collected.
	% Despite this flexibility, our FOON is not designed to deal with objects and/or states which have never been seen in source videos.
{\color{black}To further improve the performance of knowledge retrieval as a follow up to our previous work, we discuss generalizing knowledge to be applied to objects which are similar to what we have in FOON without manually annotating new sources of knowledge.
We discuss two means of generalization: 1) expanding our network through the use of object similarity to create new functional units from those we already have, and 2) compressing the functional units by object categories rather than specific objects.
We discuss experiments which compare the performance of our knowledge retrieval algorithm with both expansion and compression by categories.}
\end{abstract}

% \begin{keywords}
% knowledge representation, affordance learning
% \end{keywords}

%===============================================================================

\section{Introduction}
\label{sec:intro}
There has been a recent boon in studies regarding the importance of the theory of affordances \cite{Gibson_1977} in learning and understanding behaviour in human activities.
Studies in neuroscience and cognitive science on object affordances indicate that the mirror neurons in human brains congregate visual and motor responses \cite{Rizz_2004, Rizz_2005, Oztop_2006}.
Mirror neurons in the F5 sector of the macaque ventral pre-motor cortex fire during both observation of interacting with an object and action execution, but do not discharge in response to simply observing an object \cite{Di_1992, Gallese_2002}.
Further studies \cite{Borghi_2012} show the functional relationship between paired objects and compared it with the spatial relationship and found that both the position and functional context are important and related to the motion; however, the motor action response was faster and more accurate with the functional context than with the spatial context.
Yoon et al. \cite{Yoon_2010} recently studied affordances associated to pairs of objects positioned for action and found an interesting so-called ``paired object affordance effect'', where the response time by right-handed participants was faster if the two objects were used together, where the active (manipulated) object was to the right of the other.
% v(supposed to be manipulated)
%

% The study results in neuroscience and cognitive science indicate that there are strong connections between the observation of objects and functional motions.
From these studies, it is clear that functional relationships between objects are directly associated with motor actions.
This interesting phenomenon can be observed in human daily life: when humans are performing tasks, they not only pay attention to objects and their states but also to object interactions caused by manipulation.
The manipulation reflecting the motor response is tightly associated with both the manipulated object and the interacted object.
In robotics today, there has been a significant number of works which focus on representating and learning of object features, motions and affordances \cite{saxena2014robobrain, Konidaris, GuptaD,Kjellstrom,Gall,SunRAS2013,koppula2013learning,moldovan2012learning,stoytchev2005behavior} and modelling such relationships \cite{Aksoy,Aksoy_2,jain2009}.

Motivated by the promise in object affordance-based learning, we present a novel, graphical knowledge representation method called the {\it functional object-oriented network} (FOON) \cite{Paulius2016} which represents the relationship between objects and their associated functional motions.
A single {\it functional unit}, our basic learning unit, represents a single, atomic manipulation action (such as cutting, picking-and-placing, pouring, or shaking) by capturing the objects required for such a task and the motion occurring.
A collection of these units can be used to describe the chain of events needed for a certain outcome; such a collection referred to as a {\it task tree} can be followed in sequence by a robotic system to accomplish a given task.
% We shall discuss and demonstrate how we can perform manipulation tasks given a target goal;
% 	knowledge retrieval produces a task tree (if a solution exists) which a robot can then use to execute a plan.
This representation is similar to other representations such as Petri Nets (PN)\cite{Petri:2008}, which is also a bipartite network representing events as {\it transitions} with conditions which need to be present for these actions to occur.
FOON is specifically designed for manipulation tasks, as transitions are equivalent to motion nodes and {\it conditions} (or {\it places}) are equivalent to object nodes.
Certain objects must be present for a certain action to occur just as input places are required for a transition to fire in a PN.
% In this way, a FOON can be seen as a specialized Petri Net.

% Although this method provides flexible and novel manipulations, FOON does not work with unseen cases, as there must be knowledge on unknown objects to manipulate them.
With our current methodology, learning new actions as FOON graphs requires annotating new sources and videos by hand and appending that knowledge to the universal FOON, which for us is a very time-intensive process.
To create graphs automatically is inherently a difficult problem due to the challenge in recognizing object states and motions in 2D videos from the Internet.
% We are currently exploring means of developing a system for automatic activity recognition, as the task of watching videos and recording functional units is a time-intensive process.
{\color{black}For the time being, we explore another avenue in learning a more generalized FOON due to the complex nature of automatic object-motion recognition to improve upon \cite{Paulius2016}.}
In this paper, we investigate means of creating useful knowledge for our representation through generalization using {\it expansion} and {\it categorization} with object similarity.
The intuition behind this is as follows: if we know how to manipulate a certain set of objects, then we can also manipulate those which are similar to it; from another point-of-view, if we do not have a specific object, we can use items which are similar to it to complete the task.
We can create new forms of FOON whose knowledge extends that of a regular network using these principles: an expanded network and a compressed network based on object categories.
% The main contribution of this work is a generalized view of FOON which can theoretically perform better since we present more knowledge through expansion and compression.
We shall discuss the details of each method and discuss results of knowledge retrieval using such approaches.
% In a sense, this concept relates to \cite{Aksoy,Aksoy_2}, except we characterize objects based on another source of information instead of looking for overlap in manipulations.
% We discuss results using similarity to aid the knowledge retrieval process by building upon the knowledge that is known beforehand.
% We also explore the use of generalizing a FOON through the use of {\it object categorization} which allows us to abstract knowledge
% 	to make it transferrable to other objects which we did not have in our original FOON.
% We discuss these two measures for expanding our knowledge and applicability of our network through the use of experiments to
% 	determine whether or not we can learn anything useful without obtaining more source videos.
% We discuss the capacity for learning new concepts based on concepts known already through the use of association but at the cost of scalability issues.

% Another concept we introduce to our FOON structure is the idea of hierarchies of information which can be used for abstracting and condensing the graph.
% There may be instances where we do not care about special details about objects (looking at them abstractly) versus the need for a magnified view on objects (looking at their states or ingredient composition)
% due to the robot's limited vision system. We can create simpler versions of a FOON to make FOON generalizable for vision systems of different capabilities.

\begin{figure}[h]
	\centering
	\includegraphics[width=6cm]{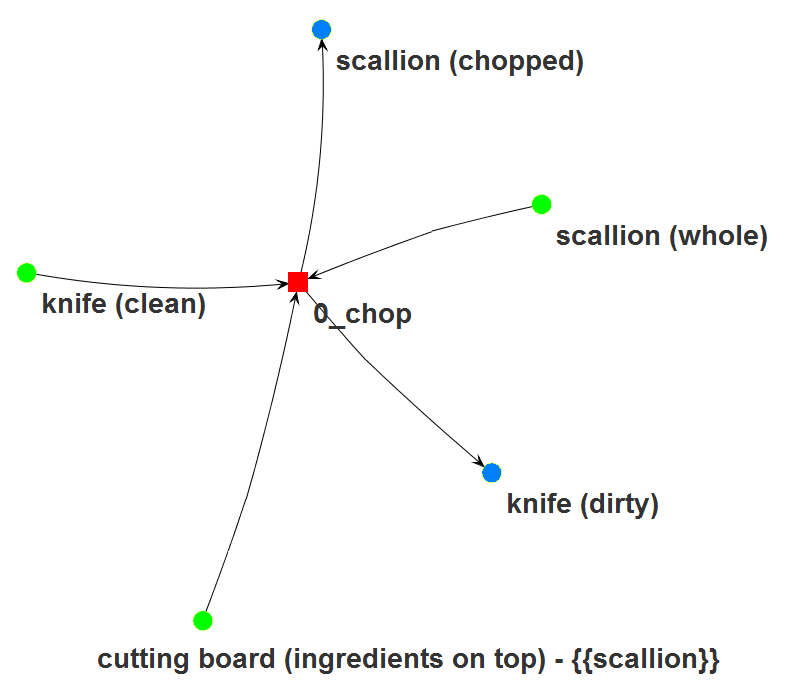}
	\caption{A basic functional unit with three input nodes (in green) and two output nodes (in blue)
		connected by an intermediary single motion node (in red). 
        Input and output nodes are distinguished by the direction of the edges.}
	\label{fig:unit}
\end{figure}

\section{Functional Object-Oriented Network}
\label{sec:network}
% We introduced a knowledge representation scheme which encodes observed activities based on observed objects and the motion they are interacting with.
% FOON therefore uses two types of nodes to capture this relationship: {\it object nodes} and {\it motion nodes}.
% These nodes are gathered as collective units which individually reflect a single action.
% These actions can belong to any domain involving object-motion manipulations, but presently our attention is placed on cooking tasks.

The {\it functional object-oriented network} is a knowledge representation which we use to encode observations on activities or functional manipulations with objects.
FOON contains information on object-motion affordances as observed in manipulation activities; in particular, it is a directed graph with two types of nodes: {\it object nodes} and {\it motion nodes}.
The purpose of this knowledge source is to capture information to be used by a robot as an executable sequence or program to solve manipulation problems (which require the use of a specific set of objects) that produce a target product.
Presently, the knowledge contained in FOON is centred around cooking activities and recipes in the kitchen, but this representation can also be extended to other manipulation-centric domains such as cleaning or factory assembly.

\subsection{Basics of a FOON}
As stated before, a FOON contains two types of nodes. 
A graph of this nature is more formally referred to as a {\it bipartite network} \cite{Newman}.
The object nodes (denoted as {\bf $N_{O}$}) in a FOON are those which are being manipulated in the environment or those which are used to manipulate other object nodes. 
Generally speaking, we solely focus on objects which are actively used or acted upon in a specific activity.
The other type of node, motion nodes (denoted as {\bf $N_{M}$}), is a node which describe the manipulation using said objects.
These motions may range from picking-and-placing, pouring, cutting or stirring.
An object node is characterized by an {\it object type}, an observed {\it object state} and a {\it motion identifier} which describes if the object is moving in the scene.
Objects can also contain other objects within them, and so we can identify objects by their {\it ingredient content}; examples of such containers include the bowl, cup and box objects.
Motion nodes are solely identified by a {\it motion type} from a set of pre-defined motions or actions as typically seen in cooking.
% As with typical bipartite networks, each node type connects to another arbitrary node type.

In a FOON, object nodes are connected to motion nodes and motion nodes are connected to object nodes.
The only instance where objects may be connected to objects is when we transform our bipartite network to a {\it one-mode projected graph} for the purpose of network analysis \cite{Newman} as we have done previously in \cite{Paulius2016}.
Edges in our graph are drawn from one node to another with respect to the order of a sequence leading to a particular object-state outcome occurring.

% \subsection{Functional Unit}
% The fundamental learning unit in a FOON is referred to as the {\it functional unit}.
% A functional unit represents a single, atomic observed action and reflects the interaction of objects via a manipulation motion.
% In a functional unit, we denote the initial state of the objects as input object nodes, the motion being carried out as a motion node and the resulting effect on the objects
% as output object nodes. As discussed before, an object may not necessarily change its state after interacting with a motion node.
% There may be functional units which intersect if there are several ways of making a specific object.

\subsection{The Functional Unit}
A FOON consists of individual, fundamental learning units for which we coined the term {\it functional units} (shown in Figure \ref{fig:unit}).
A functional unit represents a single, atomic action, and a set of functional units which are connected together form what we call a {\it subgraph} which describes an activity involving two or more steps.
For instance, in a video where the demonstrator is preparing macaroni and cheese, a subgraph is created to describe the entire activity. 
This subgraph may consist of several units describing actions such as boiling of water in a pot, stirring of macaroni pasta in a pot with a spoon, and pouring the cooked pasta into a pan.

A functional unit has three components: {\it input} object nodes, {\it output} object nodes, and an {\it intermediary motion} node which describes the cause of the transition in objects' states.
A motion does not guarantee the change in all input object states, so there may be instances where an object remains in the same state in multiple units.
Because of this, a FOON can be more accurately defined as a {\it directed acyclic graph} since there may be an incidence of loops.

\section{Creating a FOON}
A FOON is created from informational sources of knowledge, whether they be demonstrations of human activities or observations extracted from instructional videos.
However, automatic extraction of knowledge from such sources is very complex due to the difficulty in recognizing the objects being used, the states they are in, and the motion occurring.
For the time being, we annotate videos manually in lieu of such a system; volunteers were tasked with the selection and annotation of cooking videos.
% This process involves recording the objects being manipulated within the scene, noting any changes in states, and the type of manipulation motion happening.
We also note the timestamps at which actions occur in source videos for reference.
% In this section, we will be describing the steps taken in building a FOON.
% , and more recently, we have also integrated knowledge taken from the MPII Cooking Activities Dataset \cite{Schiele:2012:DFG:2354409.2354909}.

\subsection{Gathering and Combining Knowledge}
The knowledge represented in a FOON is taken from a collection of video sources found on YouTube.
A subgraph will be created for each source video, where functional units are directly constructed through manual annotation.
The annotation process simply involves the recording of actions occurring in videos, specifically the time they occur, the objects being manipulated, the changes in their states (if any), and the type of motion occurring.
% As we defined earlier, a subgraph is a collection of functional units describing a certain activity such as preparing scrambled eggs or making macaroni and cheese.
With these individual subgraphs, we can then combine the knowledge into a single, larger FOON through a merging procedure.
The merging process is very simple in intuition: we perform a union operation with all functional units while removing any duplicate units.
A duplicate in this sense means that two units have the exact same input, object and motion nodes down to the smallest detail.
Before merging, we parse these files to ensure that all labels are consistent with the object and motion indices kept as reference.
For more details on the algorithm, including pseudocode, we refer readers to \cite{Paulius2016}.

% The algorithm is as follows: we start with an initially empty functional unit list which we refer to as $G_{FOON}$.
% All functional units will be appended to this list.
% For each subgraph, we iterate through all functional units $FU_i$ and we add each of them to the list if there is no instance of such a unit in the list.
% While iterating through all subgraphs and their units, we construct and maintain an index of all object-state nodes and motion nodes to prevent any duplicates.
% The final product will be one single graph with content spanning multiple video sources.
% Because of the diverse nature of this network, we can learn and link ideas of preparing certain items differently with a wider sense of knowledge.

\subsection{Hierarchies of Information}
{\color{black}Following our previous work in \cite{Paulius2016}, we developed a new way of presenting knowledge in our graph through different levels of abstraction.
We can condense the information presented in FOON in an abstracted way through the use of {\it hierarchies}.
Here, abstraction means that we want to consider an object with as few details as possible; more specifically, we may not consider an object's state or content.}
Hierarchies are useful when we do not require as much detail for performing manipulation tasks; for instance, an object recognition system may not be built to detect certain objects in a variety of states, and so we may refer to a version of FOON which does not take states into account.
The lower the hierarchy level, the less information is given to object nodes in functional units and the fewer functional units in a FOON (from less instances of duplicate units).

We can show object nodes in three levels of abstraction:

\begin{enumerate}[label = {\bf Level \arabic*:},leftmargin=*,align=left]
\item A FOON at the {\it purely object} level 
	\begin{itemize}[leftmargin=-9mm]
		\item Objects are considered without any states or ingredient content, e.g. an object of type {\it 					``strawberry''} with state {\it ``peeled''} and another with state {\it ``chopped''} are seen as one 				object node {\it ``strawberry''}.
	\end{itemize}
\item A FOON at the {\it object-state} level 
	\begin{itemize}[leftmargin=-9mm]
		\item Objects are differentiated by their states and not ingredient content, e.g. as with the case 					above, we have two different object nodes because there are two instances with different states {\it 				``chopped''} and {\it ``peeled''}.	
		\item However, if we have two objects with mixtures like a bowl of eggs mixed with salt and a bowl of 				eggs mixed with milk, salt, and pepper, we treat them as one object node in a FOON: a bowl with {\it 			``ingredients mixed inside''}.
	\end{itemize}
\item A FOON at the {\it object-state-content} level 
	\begin{itemize}[leftmargin=-9mm]
		\item Objects in different states are classified as unique, separate nodes by considering their {\it 				composition}: what ingredients make up that object-state node.
        \end{itemize}
\end{enumerate}

\subsection{A Universal FOON}
We define a {\it universal FOON} as a merged set of two or more subgraphs from information sources.
Since a universal FOON is comprised of knowledge from multiple sources, it can be used by a robot as a knowledge base for solving problems using object-motion affordances.
At the time of this paper, our universal FOON is made up of knowledge reflecting 65 YouTube source videos covering a range of recipes.
% This FOON's size (its object nodes and motion nodes) depend on the hierarchy level used;
Table \ref{table:FOON} presents a tally of nodes at each hierarchy level.
We provide interested readers with illustrations of the FOON graphs discussed in this paper and all video subgraph files for download at our website \cite{foonet}.
% \footnote{\label{note1} \textbf{FOON Graphs and Videos} : \url{http://www.foonets.com}}.

\begin{table}[h]
	\centering
	\caption {{Universal FOON statistics at each hierarchy level.}}
	\begin{tabular}{|c|c|c|}
		\hline
		{\bf \emph {Hierarchy Level}} & {\bf \emph {\# of Object Nodes}} & {\bf \emph  {\# of Motion Nodes}} \\\hline
			Level 1	& 185	& 659	\\	\hline
			Level 2	& 911	& 866	\\	\hline
			Level 3	& 1676	& 984	\\	\hline
	\end{tabular}
\label{table:FOON}
\end{table}

% as well as the MPII Cooking Activities Dataset videos.
% \begin{figure}[t]
% 	\centering
% 	\includegraphics[width= 8cm]{result1.png}
% 	\caption{Our current level 3 universal FOON that is constructed after merging 65 instructional YouTube videos.}
% \label{fig:FOON}
% \end{figure}

\subsection{Knowledge Retrieval}
With a universal FOON, a robot will be equipped with knowledge which it can use to solve manipulation tasks given a target goal.
A human user may ask a robot to prepare a meal, given certain constraints, for example.
The aim of knowledge retrieval is to find a {\it task tree}: a sequence of functional unit-based steps which, when executed, accomplishes a goal.
A task tree is simply a collection of functional units which are likely to be connected together which, if followed in sequence, play out the execution of steps that solve a manipulation goal.
This goal can be any object node within FOON, whether it is a finished product or an intermediary-state object.
The procedure for retrieving a task tree sequence draws from the principles of fundamental graph searching algorithms; when searching, we explore depth-wise by functional unit, but we explore breadth-wise among objects in each unit.
In order to solve such problems, the robot needs to have knowledge about its domain, specifically what utensils or ingredients are in its surrounding environment, such that the system can ascertain whether or not there exists a solution in that instance.
The result of this search is either a task tree sequence being found (where a goal node is deemed as solvable and we have a functional unit sequence that produces the goal) or no tree due to a shortage of time or non-existent solution.

Our search considers the number of units (or steps) as a heuristic for finding the optimal task tree.
There may be multiple units that make an object (i.e. multiple trees with/without the same step size), but the search method considers the first unit which can be executed entirely (or specifically, where all objects required are available as input to that unit).
% \subsection{Modifying the Knowledge Retrieval}
% Due to the introduction of hierarchies in FOON, we implemented task tree retrievals in varying levels of abstractions.
% This is made possible because we keep each level separated in different structures; therefore, a search would simply require a search in a specific hierarchy level list.
% The addition of hierarchies can simplify the searching procedure, as we can potentially rely on less information on objects around us or in the execution of food preparation with a lower hierarchy level.
% As mentioned previously, a lower level of FOON will have less nodes and functional units which therefore makes the searching process quicker.
% These capabilities along with FOON's capacity to grow and combine knowledge from many sources of information allow us to obtain flexible and novel manipulation executions.
% We can therefore reduce the space for searching to make it much faster and feasible, as a robot's state recognition system will be very difficult to train for detecting a plethora of possible states or identifying very specific mixtures or ingredients.
Instead of using a step-based heuristic for finding a tree, we can also settle ties in functional units based on {\it task complexity}.
% It may be important to consider the difficulty of certain motions.
In some instances, a robot may be unable to perform a specific motion because of limitations in its configuration space or in its architecture.
However, we may possibly compensate for this by executing a simpler manipulation which gives us the same results.
As we continue to develop FOON, we would also need to make adjustments for other constraints such as preparing meals without certain ingredients to account for dietary preferences.

\begin{figure*}[t]
	\centering
	\includegraphics[width= 9 cm]{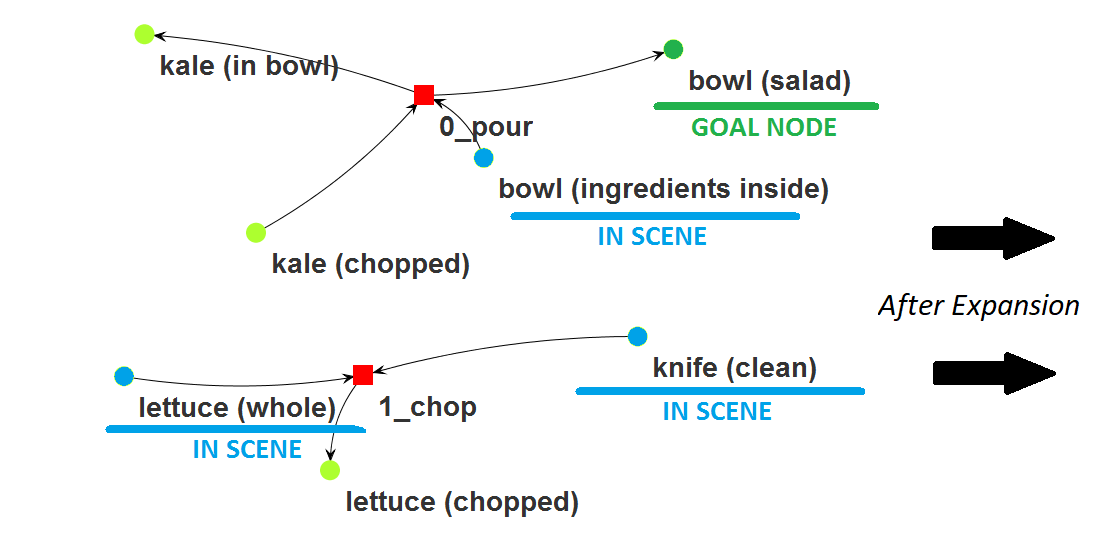}
	\includegraphics[width= 6 cm]{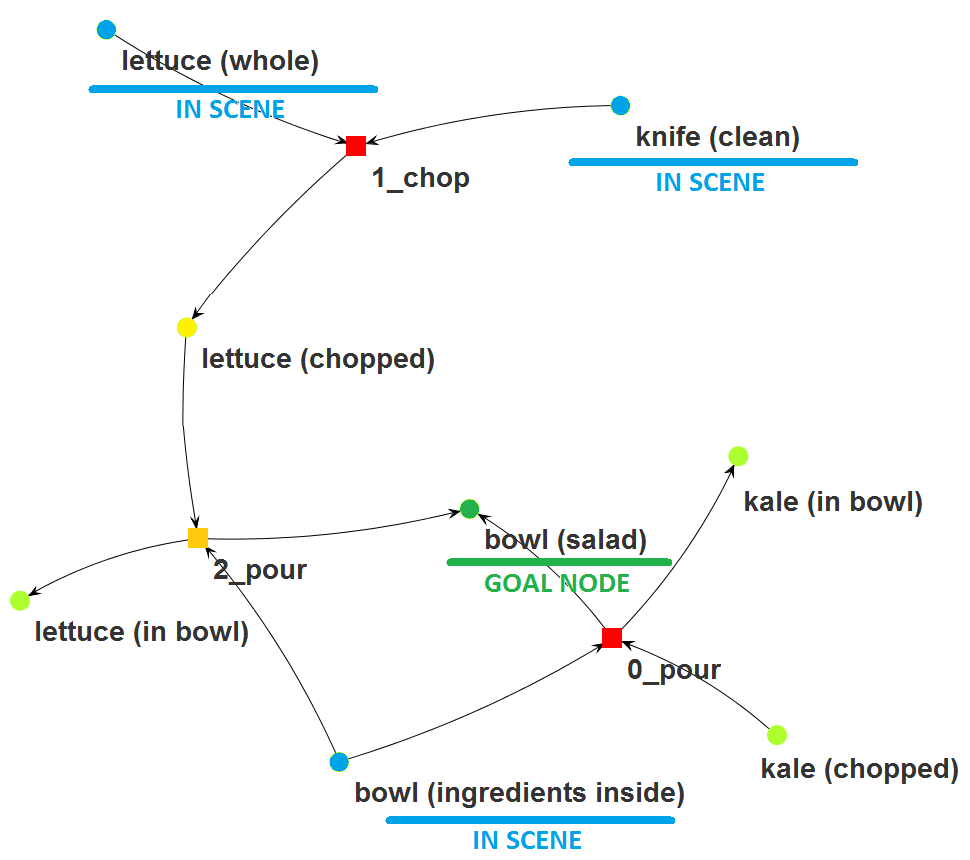}
	\caption{ An example of how expansion can be used in filling gaps of knowledge.
			Here we wish to make a salad (node in dark green) using lettuce and other items in the environment (in blue); initially, we only have knowledge on making salads with kale.
			Using similarity, we understand that kale and lettuce are similar, as they are both leafy vegetables. We create the knowledge of chopping lettuce and adding it to a bowl with other ingredients to make a salad.}
	\label{fig:expand}
\end{figure*}

\section{Generalization of FOON}
% One challenge identified is the need for additional hierarchies of generalized information.
Although FOON performs well with tasks using objects we are familiar with (i.e. preparing meals for which we have collected data as subgraphs), it is not designed to handle newly encountered objects or goals.
Sequences in FOON can only be solved as we have learned which may be an issue in adapting to heterogeneous environments.
{\color{black}For example, this suggests that if an object is not observed within a state that is necessary according to a FOON, it means that we can possibly fail to find a task tree to achieve a goal.}
Instead of quitting when we cannot solve a problem with unknown objects, we should still try to apply the knowledge contained in a FOON to complete the task.
We can use knowledge of how we use one item and transfer that knowledge to using other items like it.
This is much like our common sense as humans: if we know how to use a certain item, then we can use {\it similar} items in a similar fashion.

This would require us to create a FOON which is generalized but not like how we implemented hierarchy levels.
Rather, we need to create a FOON which is generalized to account for a wider set of cases.
We explore this through two ideas: 1) we can create an extremely large variant of FOON which contains all possible combination of uses of objects as functional units, or 2) we create a compressed variant of FOON which contains units which describes how classes or categories of objects are used. 
% which either includes functional units of every single object type or we create a FOON with functional units which only considers the categorization of objects.
This categorization can therefore extend to items within a subset and thus those which can be manipulated in the same way.
These two basic ideas motivate us to investigate two ways which can possibly improve the usability of a FOON.
% These methods shall be discussed further in this section.

\subsection{FOON Expansion with Object Similarity}
We have explored the use of {\it object similarity} for the purpose of finding relationships between objects we know how to manipulate in FOON and those for which we are missing knowledge.
One issue with our previous edition of FOON is that we will miss out on very basic steps (or functional units) because of missing information in source videos.
For example, we may see onions in the {\it ``chopped''} state on a plate, but we never see the {\it ``chop''} action using a knife object on a {\it ``whole''} onion.
However, what if we are confronted with a situation where we only have the natural state of onions available and yet we require it chopped?
In the case of the previous example, if we know how to chop something {\it similar} to onions such as chives, then we can theoretically perform the same manipulation with onions.

{\color{black}Using the intuition of object similarity, we can create new knowledge (i.e. creating new functional units) by using what we have collected as reference for new objects.
The {\it expansion} algorithm for creating a larger FOON involves copying functional units which already exist and creating new units with similar objects.
An example of how expansion can be used is shown as Figure \ref{fig:expand}.
This will be done for every combination of objects which are similar to one another.
We refer to an expanded FOON as {\it FOON-EXP}.
Statistics for the FOON-EXP used are shown in Table \ref{table:FOON-S-0.9}).}
% ; given a functional unit of chopping kale, we can create another unit of chopping lettuce for salad

\begin{table}[t]
	\centering
	\caption {{Statistics of our universal FOON after expansion (FOON-EXP) using threshold of 0.89.}}
	\begin{tabular}{|c|c|c|}
		\hline
		{\bf \emph {Hierarchy Level}} & {\bf \emph {\# of Object Nodes}} & {\bf \emph  {\# of Motion Nodes}} \\ \hline
		Level 1	& 232	& 3528	\\	\hline
		Level 2	& 1996	& 5493	\\	\hline
		Level 3	& 5306	& 6942	\\	\hline
	\end{tabular}
	\label{table:FOON-S-0.9}
\end{table}

\subsubsection*{Measuring Similarity for FOON-EXP}
To measure object similarity (which is simply how similar an object's concept is to another), we would require a knowledge base that allows us to make that measurement.
We use WordNet \cite{WordNet} as a means of calculating semantic similarity based on its lexical organization of words and terms.
Although WordNet is a remarkably impressive database, it also has its shortcomings due to the lack of certain terms and categories.
There are also some objects which are not found in WordNet (such as {\it ``corn starch''}, {\it ``muffin pan''} and {\it ``protein powder''}), and this needs to be alleviated manually with user-defined values.
For the sake of our discussion, we only use the computed values from WordNet; certain items will not have any similar objects identified.
% One must ensure that the similarity calculations are based on the correct meaning or {\it sense} of the word, as the main definition of some terms are not the same for food objects.
We calculate similarity of two objects using the Wu-Palmer metric \cite{wu1994verbs} available to users in the NLTK package \cite{bird2009natural}.
This metric produces a similarity score from 0 to 1, where 1 indicates that two items are conceptually equal.
% This property is transitive, as two objects A and B which have the same degree of similarity will also share the same similarity with a third object C.
We would simply use a similarity threshold as a cut-off point for determining when two concepts are alike; we use a threshold value of 0.89 as it produces a FOON that is not too large to handle than our regular network.
A lower threshold of even 0.88 would result in an exponentially larger FOON, much larger than what we are using in this paper.
We calculated similarity among a list of 356 objects with WordNet and used it to expand our default FOON.

\begin{figure*}[t]
\centering
	\includegraphics[width= 7.5 cm]{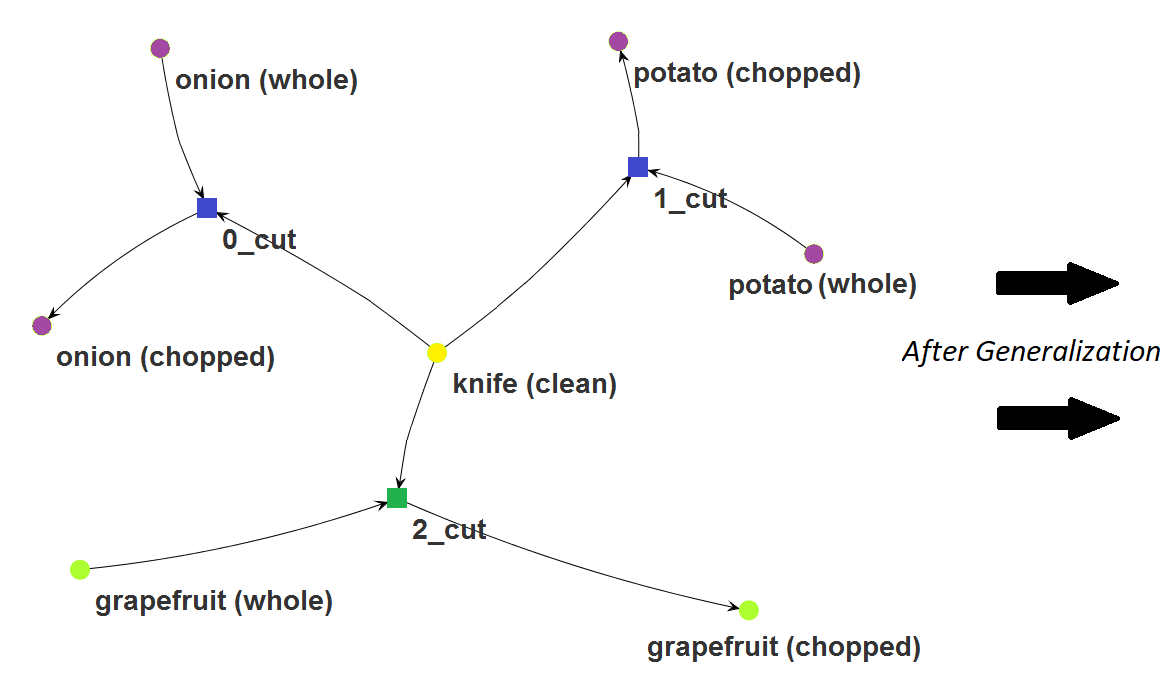}
	\includegraphics[width= 6 cm]{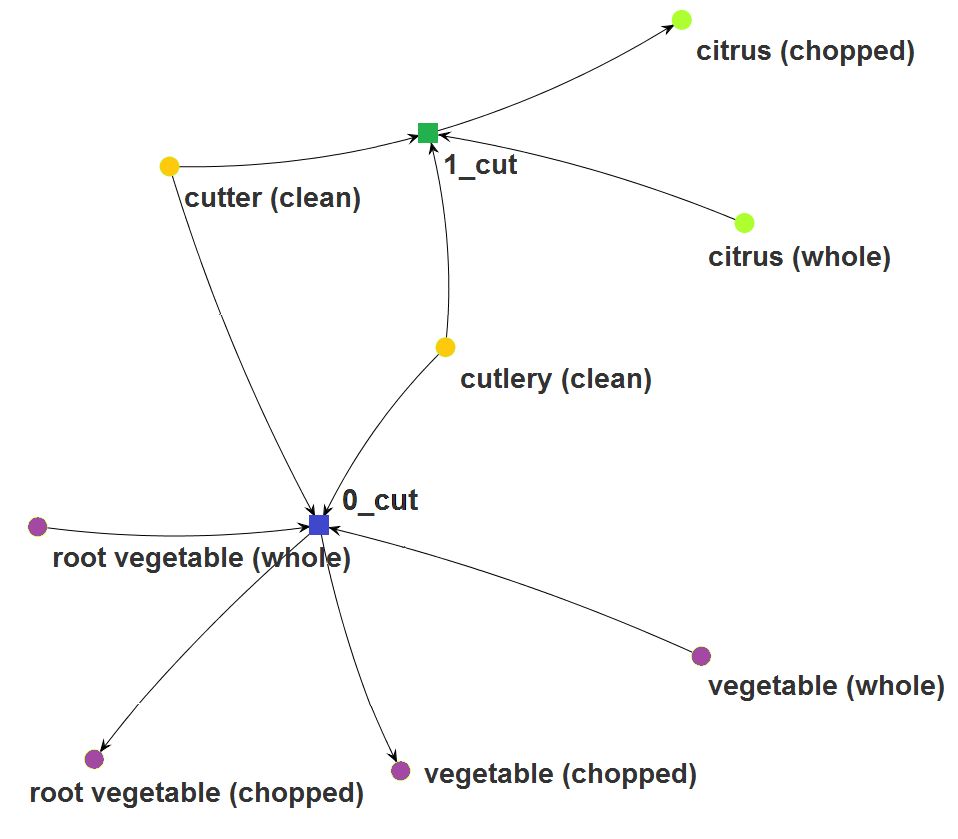}
	\caption{A set of functional units before and after generalization by categories.
			The node colours indicate which objects and categories are synonymous; the grapefruit object nodes were generalized as {\it citrus} and potatoes and onions were generalized as {\it vegetables} and {\it root vegetables}.
			Two functional units ({\it 0\_cut} and {\it 1\_cut}) end up compressed into a single unit.
		}
	\label{fig:gen}
\end{figure*}

\subsection{FOON Compression with Object Categories}
Instead of abstracting the objects using our hierarchies, we can go further by abstracting objects based on another hierarchy level, a {\it categorical} classification of objects.
In this approach, we are not concerned about specific objects but instead we focus on how {\it classes} of objects are characteristically used and manipulated.
For example, fruits share a common trait in having seeds, and so we will cut most of them expecting the seed(s) to be in the centre.
For the purpose of discussion, we will refer to the generalized version of our universal FOON as {\it FOON-GEN}.
Functional units can be represented using generic categories which can then be applied to a wider range of objects without the need for expanding our network and adding many more functional units.
This prevents the network from ``blowing up'' from a drastic increase in size from all of the functional units added from this process.
A major requirement is predefining categories for classification.
% We can also prevent unusable functional units from being created and appended to the network.
\newline

\subsubsection*{Creating a FOON-GEN}
Items can fall under one category, several categories, or possibly no categories if they are too unique an object.
We have defined a list of 56 categories; these include {\it spices}, {\it open containers}, {\it condiments}, {\it vegetables} (and further classification as {\it leafy vegetables} or {\it root vegetables}), {\it cutlery}, and {\it eating utensils}.
These categories have been defined in terms of functionality and object types, but we can still account for other functional considerations such as shape and textures.
We initially fill each category using WordNet and then we correct it manually by allocating and grouping items ourselves.
This is because WordNet lacks certain concepts as we have in our list of items, or more commonly, items were found and misclassified even through there was a similarity of object and category found.
% Once again, this is because of how WordNet itself is compiled, and so we need to compensate for this by correcting the index ourselves.
% \subsubsection*{Creating a FOON-GEN}
With the object-category index mapping 359 possible objects to 56 categories, we can then construct a FOON labelled in terms of categories.
The procedure for producing a FOON-GEN using categories is a simple process: we iterate through all of the functional units and we search for all objects which belong to a specific category, for all categories.
Once we find these objects, we can simply replace them with the category and then append these new units to a separate list.
Therefore, these functional units do not refer to a specific object but instead it will use the generalized concept of an object (once the item(s) have a mapping to the list of categories).
An example of how categorization works is shown as Figure \ref{fig:gen}.
% We expect that by using generalized functional units, we further reduce the amount of information kept in FOON for
% 	task retrieval.

The version of FOON-GEN used in our experiments is a Level 2 FOON, and it has a total of 1643 nodes comprising of 822 object-category nodes and 821 motion nodes. 
This is a smaller fraction of the number of functional units featured in our universal FOON and its FOON-EXP version.
A smaller graph would allow for faster searching times so we expect that FOON-GEN would perform well in task tree retrieval.

\section{Evaluation of FOON Generalization Methods}
In the previous section, we discussed details for methods for making our FOON more generalized for use in solving problems which we do not have knowledge of.
We have outlined two ideas: FOON expansion and compression.
In this section, we justify the usefulness of our approaches as an improvement in solving unknown problems.
We evaluate our generalization methods by measuring how likely each version of our universal FOON is to contain the knowledge needed to solve a particular problem.
We hypothesize that our generalized method using categories will outperform an expanded FOON and our default universal FOON.

\subsection{Methodology}
Our experiments were conducted as follows: during a series of 10 trials, we randomly select 100 goal nodes as target products and we attempt to find a task tree that produces each of these goals.
The network with the most successes (i.e. the highest recorded number of objects for which a task tree was found out of the total number of goal nodes) will indicate the best of the three methods.
These goal nodes are those which are not in its basic state, i.e. it must be the output of a functional unit.
In this way we do not consider items which we may be searching for which are already in our kitchen.
In each trial, we simulate different layouts of kitchen environments by randomly selecting a subset of ingredients/utensils out of a possible 224 object nodes.
We do this since our knowledge retrieval algorithm requires a list of items in the environment.
The hierarchy level (as hinted before) which we perform our experiments is in Level 2 FOONs, meaning that we only work with object and states and neglecting specific ingredient combinations.
We will be measuring each trial by the average time taken for task tree retrieval in addition to the number of objects successfully found.
An object with no task tree found within a certain period of time (or in our case, a certain number of iterations) is considered as unsolvable and has no existing solution.
This is important when considering obtaining task tree sequences in real-time as needed by robots.

\subsection{Discussion of Results}
We ran the experiments while recording average running times (shown in Table \ref{table:FOON-EX2}) and the number of objects we successfully found task trees for out of a possible 100 object goal nodes (shown in Table \ref{table:FOON-EX}).
For time complexity considerations, these experiments were run on a machine with an Intel Core i7-6500U processor and 12 GBs of RAM.

These results show that the generalized FOON performs much better than the other networks, as we were able to find at least 66\% of all object goal nodes using only a subset of kitchen ingredients and utensils among all trials.
FOON-GEN provided for the fastest searches on average, validating that a network of smaller size would require less time to search.
Therefore, a generalized representation would be ideal for solving problems in real-time.
The FOON-EXP network also does fairly well when compared to the regular network; in most instances, it outperforms the regular network.
Theoretically, a FOON-EXP should allow for at least as many as the regular network, as the regular network is a subset of the expanded network.
However, the expansion dramatically increased the number of functional units in the network, and so the searches will require more time to find a task tree.
In trial 10, the regular network outperformed FOON-EXP, as task trees for 4 more objects were found.

\begin{table}[t]
	\centering
	\caption{Average running times over all trials of random searches with unexpanded network (REG),
			expanded network using object similarity (EXP), and
				a generalized network with categories (GEN).
					% Underlined is the lowest average time, which was observed in task tree retrievals in the generalized network.
			}
	\begin{tabular}{|c|c|}
		\hline
		{\parbox{2cm}{\centering \bf \emph {FOON Selected}}} &  {\parbox{4cm}{\centering \bf \emph {Average Time Over Trials (ms)}}} \\ \hline
 		REG & 2120.5				\\ \hline
		EXP & 15217.4				\\ \hline
		GEN & {\underline{1167.5}} 	\\ \hline
	\end{tabular}
\label{table:FOON-EX2}
\end{table}

\begin{table}[t]
	\centering
	\caption{Results of random-search experiment with an unexpanded network (REG), an expanded network using object similarity (EXP), and a generalized network (GEN).
			% To clarify, for example, we observed in trial 4 of our experiment that the expanded network (EXP) performed better than the regular network (REG),
			% 	as we successfully found task trees for 33 out of 50 goal nodes as opposed to 30 objects.
			% The generalized network (GEN) in this same trial outperforms the other two, as task trees for 42 out of 50 goal nodes were found.
			}
	\begin{tabular}{|c|c|c|c|c|c|c|c|c|c|c|}
		\hline
		\multirow{2}{*}{\parbox{1cm}{\centering \bf \emph {FOON Selected}}} & \multicolumn{10}{c|}{\bf \emph {Trials for Experiment}} \\ \cline{2-11}
		& {\bf \emph {\#1}} & {\bf \emph {\#2}} & {\bf \emph {\#3}} & {\bf \emph {\#4}} & {\bf \emph {\#5}} & {\bf \emph {\#6}} & {\bf \emph {\#7}} & {\bf \emph {\#8}} & {\bf \emph {\#9}} & {\bf \emph {\#10}} \\ \hline
 		REG & 53 & 15 & 28 & 42 & 17 & 48 & 19 & 42 & 53 & 18 \\	\hline
		EXP & 55 & 23 & 32 & 53 & 28 & 54 & 27 & 47 & 62 & 14	\\	\hline
		GEN & {\underline{66}} & {\underline{70}} & {\underline{73}} & {\underline{76}} & {\underline{78}} & {\underline{87}} & {\underline{81}} & {\underline{70}} & {\underline{78}} & {\underline{77}}	\\	\hline
	\end{tabular}
\label{table:FOON-EX}
\end{table}

\subsection{Limitations of FOON-EXP \& FOON-GEN}
The drawback to using an expanded network like FOON-EXP is that it does not perform significantly better than the regular universal FOON in all cases.
% A FOON-EXP theoretically should perform at least as well as the regular network.
By adding many more  functional units, we require a deeper and lengthier search (as suggested by the great difference in average search times in Table \ref{table:FOON-EX2}).
% We believe this is due to the addition of unusable functional units being created that do not accurately reflect the reality of how certain items can be used.
% A generalized approach like FOON-GEN alleviates this issue with fewer nodes for faster search times.
We have shown that a generalized FOON allows for more successful searches, as we alleviate the issue where expansion does not create the necessary functional units to solve a problem.
Additionally, there may be certain similarities which were suggested by WordNet which do not exactly match up to the manually defined categories to create a FOON-GEN.
However, despite this fact, certain issues remain when it comes to mapping these generalized task trees to the robot's planning of action.
As we mentioned before, this representation is only symbolic at this point, and so we need to develop a means of mapping this knowledge to a manipulation planning system.
% If we have a task tree with functional units in categories, it means that we can find all sorts of combinations of actions and item substitutes.

{\color{black}For this generalization to work effectively, categories must be defined more clearly, taking into account features such as functionality, textures, shapes, and composition.}
For example, can we use scissors in place of a knife if that's the only ``cutter'' object available?
Obviously, this would depend on what we are cutting, as scissors cannot be used for getting very clean slices of fruit, but it can be used just for cutting things into chunks.
How do we plan around those sort of situations?
Additionally, there should perhaps be some sort of internal ranking among categories, where certain items should be taken in preference over others.
For example, a knife can be used for pressing things using its flat, blunt side, but it is more likely to be used for cutting.
Such considerations need to be addressed based on these extra functional details.
For future work, we will work on creating categories for different criteria which can be adapted to the robot's needs.
{\color{black}Such features like shapes are well suited for identifying objects, and so it would be ideal to combine this theoretical knowledge with other modalities of information (3D models, feature detectors, etc.).}
% We can also consider a more specific (or less general) FOON if we require a more specialized solution to our problem.
% \begin{table}[t]
% \centering
% \caption{Count of }
% \begin{tabular}{|c|c|c|c|c|c|c|c|c|c|c|}
% \hline
% \multirow{2}{*}{\parbox{1cm}{\centering \bf \emph {}}} & \multicolumn{10}{c|}{\bf \emph {Trials for Experiment}} \\ \cline{2-11}
% 		& {\bf \emph {\#1}} & {\bf \emph {\#2}} & {\bf \emph {\#3}} & {\bf \emph {\#4}} & {\bf \emph {\#5}} & {\bf \emph {\#6}} & {\bf \emph {\#7}} & {\bf \emph {\#8}} & {\bf \emph {\#9}} & {\bf \emph {\#10}} \\ \hline

% Correct & 55 & 48 & 62 & 34 & 32 & 30 & 54 & 38 & 38 & 55	\\	\hline

% TOTAL & {\bf{66}} & {\bf{70}} & {\bf{73}} & {\bf{76}} & {\bf{78}} & {\bf{87}} & {\bf{81}} & {\bf{70}} & {\bf{78}} & {\bf{77}}	\\	\hline
% 	\end{tabular}
% \label{table:FOON-EX}
% \end{table}

\section{Conclusion and Future Work}
To conclude, we discussed the basics of FOON (from functional units to the creation of a universal FOON) and how we can extract useful knowledge as task trees through the process of knowledge retrieval.
{\color{black}Originally proposed in \cite{Paulius2016}, this knowledge representation has been introduced as a means of representing manipulations in cooking activities and household tasks in general.}
Even with the novelty of task trees produced by task tree retrieval, we are limited to the knowledge we obtain from instructional videos; if we do not see an important step in them, we miss out on valuable information which can hinder the performance of the network.
We also do not know how we can handle cases where we may want to use different ingredients.
To improve FOON, we remedy this by generalizing our FOON through two means: using {\it object similarity} to expand our network and using {\it categories of objects} to compress our network.
With the first method, we expand our network by adding new functional units based on those we have seen already, thus creating units for all combinations of similar objects.
The second method condenses the network to a generalized state by substituting objects with categorical concepts.
This network will contain less functional units than the regular network, while the expanded network will likely contain more functional units than the other representations.

In our experiments, we algorithmically compared our two methods of generalization with three versions of our FOON: an expanded version {\it FOON-EXP}, a condensed, generalized version {\it FOON-GEN}, and our untouched, collected network {\it FOON-REG}.
We showed that using a smaller, condensed network performs best as long as we have a sufficiently designed category classification.
This condensed network, {\it FOON-GEN}, also had a generally lower average time of execution due to its smaller size, making it ideal for quicker searches.
However, for this network to be usable, more developments would be required in the actual manipulation planning of the robot, as we would need to de-generalize and specify how objects are to be used to solve a problem.
We would need to build a system which goes beyond the symbolism of our knowledge representation to how the robot detects and interacts with items.
% We are therefore developing a robotic system to illustrate how a robot can use FOON for its operation and then investigate how we can implement generalization techniques to improve the quality of its task executions (either by proposing alternative.

In our future work, we aim to explore event recognition for annotating new video sources of information.
We can use probabilities based on what we already have in FOON or those reflecting other datasets and apply them to a system which can be used for identifying objects in a scene and/or the action taking place.
This would be followed by training a detection system using deep learning techniques which has the capability of detecting activities and recognizing objects used in that activity.
In this way, we can perform semi-automatic collection of new functional units which can then be appended to our FOON for further development.
Furthermore, although we have shown that FOON is theoretically useful as a knowledge source, we will perform demonstrations using a real robot to investigate scenarios in which similarity would be useful or not.
These searches are purely algorithmic, and for them to work, we would require more details and precisely defined categories.
This would require us to develop a modular system for translating the knowledge from FOON into an executable program.
{\color{black}Through these experiments, we can ascertain whether or not the hierarchies and methods proposed are sufficient for other task domains other than cooking.}

\section*{Acknowledgements}
This material is based upon work supported by the National Science Foundation under Grant No. 1421418.

% \nocite{*} - include all references, even with no citations

\bibliographystyle{unsrt}
\bibliography{ref}

% \begin{figure}[h!]
% 	\centering
% 	\includegraphics[width=1.9cm]{link.jpg}
% 	\caption*{\bf{Link to supplementary material, located at \url{https://drive.google.com/drive/folders/0BzHhGNEnIgUIamJ1YTNpTzlXWEU?usp=sharing}}}
% \end{figure}

\end{document}